\documentclass[11pt, onecolumn]{IEEEtran}
\usepackage[letterpaper, left=1in, right=1in, bottom=1in, top=1in]{geometry}
\newcommand{\eps}{\varepsilon}
\usepackage[usenames,dvipsnames]{xcolor}
\usepackage{dsfont}
\usepackage{algorithm}
\usepackage{algpseudocode}
\usepackage{amsbsy}
\usepackage{amssymb}
\usepackage{amsmath}%
\usepackage{amsfonts}%
\usepackage{amsthm, bm}
\usepackage{breqn}
\usepackage{graphicx}
\usepackage{colonequals}
\usepackage{psfrag}
\usepackage{mathrsfs}
\usepackage{mdframed}
\usepackage{enumitem}
\usepackage{mathtools}
\usepackage{wrapfig}
\usepackage{subcaption}
\usepackage{setspace}
\usepackage{booktabs}
\usepackage{multirow}
\usepackage{comment}

\allowdisplaybreaks

\makeatletter
\newtheorem*{rep@theorem}{\rep@title}
\newcommand{\newreptheorem}[2]{%
\newenvironment{rep#1}[1]{%
 \def\rep@title{#2 \ref{##1}}%
 \begin{rep@theorem}}%
 {\end{rep@theorem}}}
\makeatother
\newreptheorem{conj}{Conjecture}

\theoremstyle{definition}
 
\newtheorem{defn}{Definition}

\newcommand{\calX}{\mathcal{X}}

\newcommand{\calD}{\mathcal{D}}

\newcommand{\calS}{\mathcal{S}}

\newcommand{\calP}{\mathcal{P}}

\newcommand{\sE}{\mathsf{E}}
\newcommand{\sK}{\mathsf{K}}
\newcommand{\sM}{\mathsf{M}}

\newcommand{\calW}{\mathcal{W}}

\definecolor{light-gray}{gray}{.90}
\definecolor{aliceblue}{rgb}{0.94, 0.97, 1.0}
\definecolor{airforceblue}{rgb}{0.36, 0.54, 0.66}
\definecolor{bleudefrance}{rgb}{0.19, 0.55, 0.91}
\definecolor{cerulean}{rgb}{0.0, 0.48, 0.65}

\newmdenv[%
  linewidth =0pt,%
  fontcolor=bleudefrance,
  innertopmargin=2pt,
  innertopmargin=2pt,
  leftmargin = 0pt,
  rightmargin = 0pt,
  innerleftmargin = 2pt,
  innerrightmargin = 2pt,
  skipabove = 6pt,%
  skipbelow = 6pt
]{RQ}

\newmdenv[%
  backgroundcolor=aliceblue, %
  linewidth = 2pt,%
  skipabove = 10pt,%
  skipbelow = 10pt,
  pstrickssetting={linestyle=dashed,},
  linecolor=airforceblue,
  middlelinewidth=2pt
]{TODO}

\usepackage{setspace}
\usepackage{hyperref}
\author{}
\date{}
\newtheorem{lemma}{Lemma}
\newtheorem{proposition}{Proposition}
\newtheorem{theorem}{Theorem}
\newtheorem*{theorem*}{Main Theorem}

\begin{document}

\title{Privacy Loss of Noisy Stochastic Gradient Descent Might Converge Even for Non-Convex Losses}

\author{%
Shahab Asoodeh and Mario Diaz\thanks{S. Asoodeh is with Department of Computing and Software, McMaster University (e-mail: asoodeh@mcmaster.ca). M.~Diaz is with the Instituto de Investigaciones en Matem\'{a}ticas Aplicadas y en Sistemas, Universidad Nacional Aut\'{o}noma de M\'{e}xico, Coyoac\'{a}n, 04510, Mexico (e-mail: mario.diaz@sigma.iimas.unam.mx).}
}

\maketitle

\begin{abstract}
The Noisy-SGD algorithm is widely used for privately training machine learning models. Traditional privacy analyses of this algorithm assume that the internal state is publicly revealed, resulting in privacy loss bounds that increase indefinitely with the number of iterations. However, recent findings have shown that if the internal state remains hidden, then the privacy loss might remain bounded. Nevertheless, this remarkable result heavily relies on the assumption of (strong) convexity of the loss function. It remains an important open problem to further relax this condition while proving similar convergent upper bounds on the privacy loss. In this work, we address this problem for DP-SGD, a popular variant of Noisy-SGD that incorporates gradient clipping to limit the impact of individual samples on the training process. Our findings demonstrate that the privacy loss of projected DP-SGD converges exponentially fast, without requiring convexity or smoothness assumptions on the loss function. In addition, we analyze the privacy loss of regularized (unprojected) DP-SGD. To obtain these results, we directly analyze the hockey-stick divergence between coupled stochastic processes by relying on non-linear data processing inequalities.
\end{abstract}

\section{Introduction}

Nowadays, Machine Learning (ML) models are trained using vast amounts of sensitive data, which increases the risk of privacy breaches \cite{Carlini, Carlini_SecretSharer}. To safeguard individuals' privacy, a commonly employed approach involves adding noise to the model parameters at each step of the training process. One popular method is the noisy stochastic gradient descent algorithm (\textsf{Noisy-SGD}), wherein Gaussian noise with calibrated covariance is added to the gradient of the batch loss function \cite{abadi2016deep}. This algorithm closely resembles non-private stochastic gradient descent and can be utilized in a wide range of machine learning architectures. Consequently, it has been integrated into mainstream ML platforms like TensorFlow \cite{TensorFlow_Privacy} and PyTorch \cite{Opacus}. However, despite the widespread usage and simplicity of \textsf{Noisy-SGD}, a tight analysis of its privacy loss, i.e., its differential privacy parameters, has remained open.

Traditional privacy analyses of \textsf{Noisy-SGD} make assumptions about the release of intermediate updates and employ composition theorems for approximate differential privacy (DP) or R\'{e}nyi differential privacy (RDP) to estimate the privacy loss \cite{abadi2016deep, Balle_AnalyticMomentAccountant}. In general, these analyses yield privacy loss bounds that grow  \textit{unboundedly} as the number of iterations increases. According to those results, in order to preserve privacy, it becomes necessary to restrict the number of iterations allowed for \textsf{Noisy-SGD}, resulting in a sub-optimal optimization error.

In many practical scenarios, only the final parameters of the model are made public. This scenario, known as the \textit{hidden state} setting\footnote{This setting was first introduced in \cite{Feldman2018PrivacyAB} and is sometimes referred to as the privacy amplification by iteration.}, prompted Chourasia \textit{et al.} \cite{chourasia2021differential} to propose a novel framework for analyzing the privacy loss of iterative algorithms. By relying on a discretization of the Stochastic Gradient Langevin Dynamic (SGLD), they derived privacy loss bounds for gradient descent that converge as the number of iterations goes to infinity. Although this result is specific to \textit{full batch} gradient descent under \textit{smooth} and \textit{strongly convex} losses, it is remarkable because it demonstrates that in the hidden state setting, it is possible to run the algorithm without necessarily more privacy degradation.

In recent studies, Ye and Shokri \cite{Shokri_SGD_SGLD} as well as Ryffel \textit{et al.}\ \cite{ryffel2022differential} have achieved significant progress in understanding the privacy guarantees in the more practical \textit{mini-batch} setting, i.e., \textsf{Noisy-SGD}. Specifically, they established convergent privacy loss bounds for \textsf{Noisy-SGD} under \textit{strongly convex} loss functions. These results represent an important step towards the understanding of the privacy guarantees of \textsf{Noisy-SGD}. Concurrently, Altschuler and Talwar \cite{Altschuler_Talwar} employed a fundamentally distinct technique to demonstrate the convergence of the privacy loss of \textit{projected} \textsf{Noisy-SGD} under \textit{convex} losses. Although the shift from strongly convex to convex loss might appear insignificant, the technical challenges involved are non-trivial and represent an important achievement in this field.

The works mentioned earlier heavily rely on the (strong) convexity of the loss function. Therefore, for instance, Altschuler and Talwar \cite{Altschuler_Talwar} acknowledged that their techniques cannot be directly applied to general \textsf{DP-SGD}, a popular variant of \textsf{Noisy-SGD} where gradients are clipped to limit the impact of individual samples on the training process. Given the widespread use of \textsf{DP-SGD} in practice, it is crucial to develop appropriate techniques that can handle the non-convex nature of this variant of \textsf{Noisy-SGD}. Thus, as pointed out in both \cite{Shokri_SGD_SGLD} and \cite{Altschuler_Talwar}, it remains an important open problem to relax the convexity (and smoothness) assumption, while proving convergent privacy upper bounds.
%as pointed out in \cite{Altschuler_Talwar}, ``convexity is too restrictive when training deep networks, and it is an interesting
%open question if the assumption of convexity can be relaxed.''\footnote{Also, Ye and Shokri \cite{Shokri_SGD_SGLD} pointed out that ``It remains an important open problem to further relax the convexity and smoothness conditions for proving converging R\'enyi DP bound.''} 
%Furthermore, Altschuler and Talwar \cite{Altschuler_Talwar} acknowledged that due to the lack of convexity, their techniques cannot be directly applied to \textsf{DP-SGD}, a popular variant of \textsf{Noisy-SGD} where gradients are clipped to limit the impact of individual samples on the training process. 
 This paper aims to address this challenge by providing bounds for the privacy loss of projected \textsf{DP-SGD} that converge exponentially fast as the number of iterations goes to infinity for \textit{non-convex} loss functions. Additionally, we present privacy loss bounds for regularized (unprojected) \textsf{DP-SGD} that exhibit favorable numerical behavior. To the best of our knowledge, this is the first study to establish convergent privacy loss bounds for a non-convex variation of \textsf{Noisy-SGD}.
\begin{figure}[t]
 \hspace{-15pt}
   \begin{subfigure}{.4\textwidth}
  \includegraphics[scale=0.4]{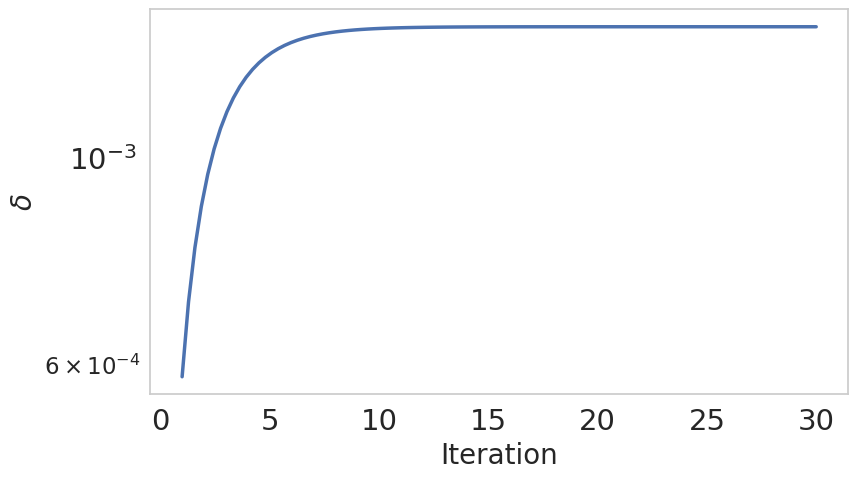}
  %\caption{$\eps = 3$}
  \end{subfigure}
  \qquad \qquad\qquad\begin{subfigure}{.4\textwidth}
  \includegraphics[scale=0.4]{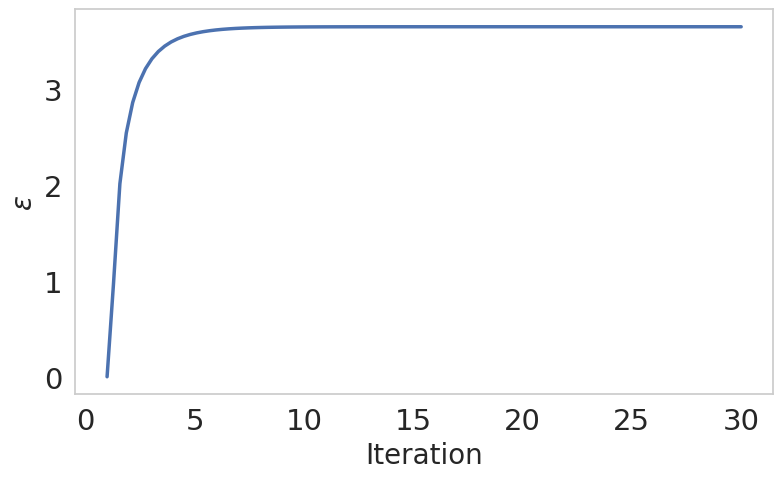}
\end{subfigure}
\caption{The privacy loss of \textsf{DP-SGD} with the  parameters: $D = 3, C = 2, \eta = 0.01, \sigma = 1, p = 0.001$, and $\eps =3$ (left) and $\delta = 10^{-3}$ (right).}
\label{fig:DP_SGD}
\end{figure}
\subsection{Our Contributions and Techniques}

Throughout, we focus on the projected version of \textsf{DP-SGD} that we introduce next. 
Let $\sigma\geq0$ and $\mathcal{W}\subset\mathbb{R}^{d}$. Given a dataset $\mathcal{X} = \{x_{1},\ldots,x_{n}\}$, we consider the iterative process $\{W_{t}\}_{t \leq T}$ determined by $W_{0} \sim \mu_{0}$ and
\begin{equation}
\label{eq:IntroDefDPSGD}
    W_{t} = \Pi\big[\psi_{B_{t}}(W_{t-1}) + \sigma Z_{t}\big],
\end{equation}
where $\Pi:\mathbb{R}^{d}\to\mathcal{W}$ is the projection onto $\mathcal{W}$, $Z_{t}$ is a standard Gaussian vector, $B_{t}$ is a random batch, and $\psi_{B}:\mathbb{R}^{d}\to\mathbb{R}^{d}$ for $B\subseteq \{1, \dots, n\}$ is the update function given by
\begin{equation*}
    \psi_{B}(w) = w - \frac{\eta}{\lvert B \rvert} \sum_{i\in B} \mathrm{Clip}\big(\nabla\ell(w,x_{i})\big).
\end{equation*}
Here, $\ell$ is the loss function, $\eta$ the learning rate, and the mapping $\mathrm{Clip}:\mathbb{R}^d\to \mathbb{R}^d$ is defined as
\begin{equation}
\label{eq:DefClipping}
    \mathrm{Clip}(v) \coloneqq \min\Big\{1, \frac{C}{\|v\|}\Big\} v,
\end{equation}
where $C>0$ is the so-called \textit{clipping constant}. In this work, we consider two batching strategies: Poisson sampling where each individual sample is chosen with probability $p\in(0,1)$, and sampling without replacement where we chose a random batch of size $b\leq n$.

For ease of notation, we let $\sK_{t} \coloneqq P_{W_{t} \vert W_{t-1}}$ be the update kernel that characterizes the dynamic of the iterative process \eqref{eq:IntroDefDPSGD}. Under the \emph{hidden state assumption}, the iterative process \eqref{eq:IntroDefDPSGD} is $(\eps, \delta)$-differentially private after $T$ iterations if
\begin{equation}
\label{eq:IntroHSDivergenceWt}
    \sE_\eps(\sK_T\cdots\sK_1\mu_0\|\sK'_T\cdots\sK'_1\mu_0)\leq \delta,
\end{equation}
where $\sK_1, \dots, \sK_T$ and $\sK'_1, \dots, \sK'_T$ are the update kernels in \eqref{eq:IntroDefDPSGD} corresponding to neighboring datasets $\calX$ and $\calX'$, respectively, and $\sE_\eps(\cdot\|\cdot)$ denotes the hockey-stick divergence. See Sec.~\ref{Section:Preliminaries} for preliminaries on differential privacy, hockey-stick divergence, and contraction of Markov kernels. Hence, to obtain a privacy loss bound for the iterative process described in \eqref{eq:IntroDefDPSGD}, it is enough to study the hockey-stick divergence in \eqref{eq:IntroHSDivergenceWt}. We do so by establishing a coupled non-linear data processing inequality, i.e., we show that there exist $\alpha_t,\beta_t\in(0,1)$ such that, for all probability measures $\mu$ and $\nu$ on $\calW$,
\begin{equation}\label{eq:IntroCoupledNonlinearDPI}
   \sE_\eps(\sK_{t} \mu \Vert \sK_{t}' \nu) \leq \alpha_t \sE_\eps(\mu \Vert \nu) + \beta_t. 
\end{equation}
% \begin{equation}
% \label{eq:IntroCoupledNonlinearDPI}
%     \phantom{\forall \mu,\nu\in\mathcal{P}(\mathcal{W_{t-1}}),} \qquad \sE_\eps(\sK_{t} \mu \Vert \sK_{t}' \nu) \leq \alpha_t \sE_\eps(\mu \Vert \nu) + \beta_t. %\qquad \forall \mu,\nu\in\mathcal{P}(\mathcal{W}),
% \end{equation}
The previous inequality is a hockey-stick divergence analogue of a recursion established by Ye and Shokri to study the R\'{e}nyi divergence associated with \textsf{Noisy-SGD} in \cite[Lemma~3.2]{Shokri_SGD_SGLD}. The proof of the coupled non-linear data processing inequality in \eqref{eq:IntroCoupledNonlinearDPI} relies on the joint-convexity of the hockey-stick divergence and the strong data processing inequality for the input-constrained Gaussian kernel proved by Asoodeh \textit{et al.} in \cite{asoodeh2020online}. In fact, note that the update kernel $\sK_{t}$ could be expressed as a composition of three Markov kernels, namely, an operator specified by the update function and the batching strategy (sub-sampling), a Gaussian kernel, and finally a projection operator. The non-linear term in \eqref{eq:IntroCoupledNonlinearDPI} comes mainly from the sub-sampling operator, while the projection operator ensures that the input for the Gaussian kernel remains bounded.

Upon the coupled non-linear data processing inequality \eqref{eq:IntroCoupledNonlinearDPI}, our main result provides a convergent upper bound for the privacy loss of \textsf{DP-SGD}.

\begin{theorem*}[Informal]
\label{Them_informal}
Let $\eps\geq0$. After $T\in\mathbb{N}$ iterations, the projected \textsf{DP-SGD} under sampling without replacement is $(\eps,\delta)$-differentially private with
\begin{equation}
\label{eq:IntroMainResult}
    \delta \leq \frac{1-[(1-p)\theta]^{T}}{1-(1-p)\theta} p\theta,
\end{equation}
where $p \coloneqq b/n$ and $\theta\in(0,1)$ is a constant depending on the privacy parameter $\eps$, diameter of the parameter space $\mathcal{W}$, the learning rate $\eta$, the clipping constant $C$, and the noise variance $\sigma^{2}$.
\end{theorem*}

While this theorem provides a convergent upper bound on $\delta$ for a given $\eps$, we can convert it to obtain a convergent upper bound on $\eps$ for a given $\delta$, as illustrated in Fig.~\ref{fig:DP_SGD}. The details of the conversion process are given in Appendix~\ref{Appendix:ConversionPrivacyLossBounds}. Note that the upper bound in \eqref{eq:IntroMainResult} is always less than 1. Therefore, our main theorem establishes, for the first time, that the privacy loss of projected \textsf{DP-SGD} converges exponentially fast without convexity or smoothness assumptions on the loss function. 

In Sec.~\ref{Section:MainResults}, we derive a privacy loss bound similar to \eqref{eq:IntroMainResult} for projected \textsf{DP-SGD} under Poisson sampling. Additionally, we investigate a regularized (unprojected) version of \textsf{DP-SGD}. While the analysis and resulting privacy loss bound for the latter are more complex, they demonstrate favorable numerical behavior. Finally, we provide a discussion of our findings and highlight open problems for future directions in Section~\ref{Section:Discussion}.

\subsection{Comparison with Previous Work}

Our work is closely connected to the rapidly expanding literature on privacy amplification, see, e.g., \cite{Feldman2018PrivacyAB,Balle2019mixing,Feldman_DP_SCO, Asoodeh_Contraction_ISIT}. Recently, Ye and Shokri \cite{Shokri_SGD_SGLD} and Altschuler and Talwar \cite{Altschuler_Talwar} established convergent privacy loss bounds for certain variants of \textsf{Noisy-SGD}. In this work, we focus on establishing convergent privacy loss bounds for \textsf{DP-SGD}, a widely used variant of \textsf{Noisy-SGD} which defines a non-convex optimization problem. A distinctive feature of our work is that, unlike previous studies in privacy amplification that rely on R\'{e}nyi differential privacy and employ RDP-to-DP conversion, we directly employ the hockey-stick divergence, enabling us to formulate bounds directly in terms of $(\eps, \delta)$-differential privacy. This is a convenient feature of our bounds, as the RDP-to-DP conversion cannot be universally applied due to restrictions on the order of the R\'enyi divergence and on the noise variance, see e.g., \cite[Remark 1.4]{Altschuler_Talwar}. Moreover, it is known that RDP is only a \textit{lossy} representation of privacy loss \cite{zhu2022optimal}. 

In our work, we study sub-sampling using a mixture of distributions and derive recursions for computing the divergence between these mixtures, akin to the approaches employed by Chourasia \textit{et al.} \cite{chourasia2021differential} and Ye and Shokri \cite{Shokri_SGD_SGLD}. However, we take a different approach by leveraging the joint convexity of the hockey-stick divergence instead of relying on the exponentiated R\'{e}nyi divergence. The resulting recursion, which we refer to as the \textit{coupled non-linear data processing inequality}, can be seen as an analog of the privacy amplification by sub-sampling framework proposed by Ye and Shokri in \cite{Shokri_SGD_SGLD}, but applied within the context of the hockey-stick divergence. The term \textit{coupled} carries a similar meaning as in the work by Altschuler and Talwar \cite{Altschuler_Talwar}, where two identical iterative processes are coupled through a potentially different dataset or initialization. In our analysis of regularized \textsf{DP-SGD}, we extend this argument by considering couplings between different iterative processes.

An essential aspect of Altschuler and Talwar's analysis involves investigating the privacy amplification achieved through the composition of projection operators with noisy gradients. In the context of the hockey-stick divergence, Asoodeh et al. \cite{asoodeh2020online} utilized the contraction coefficient of Markov kernels to derive privacy amplification bounds that take into account the diameter of the parameter space. Their work, along with subsequent studies \cite{Contraction_PNSGD_Dong, Asoodeh_Federated}, relies solely on the strong data processing inequality (SDPI) to analyze the privacy guarantees of general iterative processes. However, the sub-sampling framework does not align well with the \textit{multiplicative} nature of the SDPI. In our work, we build upon their findings and establish a coupled \textit{non-linear} data processing inequality specifically tailored to the sub-sampling framework.

An important characteristic of \textsf{DP-SGD} is the gradient clipping. As noted by Altschuler and Talwar \cite{Altschuler_Talwar}, ``clipped gradients do not correspond to gradients of a convex loss'', and hence their results (as well as all other works in the literature that aim at proving convergent privacy bounds) do not apply. To deal with the non-convexity of \textsf{DP-SGD}, we adopt an alternative approach that avoids requiring the geometric contraction of the update mappings. Instead, we solely rely on the probabilistic contraction of the Markov kernels under hockey-stick divergence. This enables us to analyze the privacy loss without imposing any convexity assumption on the loss function.

\subsection{Notation}
We use upper-care letters (e.g., $W$) to denote random variables and  calligraphic letters to represent their alphabets (e.g., $\calW$). The set of all distributions on $\calW$ is denoted by $\calP(\calW)$. 
A Markov kernel $\sK:\calD\to \calP(\calW)$ is specified by a collection of distributions $\{\sK(x)\in \calP(\calW):x\in \calD\}$. 
Given a Markov kernel $\sK:\calD\to \calP(\calW)$ and $\mu\in \calP(\calD)$, we denote by $\sK\mu$ the push-forward of $\mu$ under $\sK$, i.e., the output distribution of $\sK$ when the input is distributed according to $\mu$, and is given by $\sK\mu\coloneqq \int\mu(\text{d}x)\sK(x)$.
Notice that any deterministic function $f:\mathcal{W}\to\mathcal{W}$ can be viewed as a Markov kernel determined by $w \mapsto \delta_{f(w)}.$ Thus, $f\mu$ denotes the output distribution of $f(W)$ when $W\sim \mu$. 
We use $\mathbb{E}_\mu[\cdot]$ to write the expectation with respect to $\mu$. For an integer $n\geq 1$, we let $[n] \coloneqq \{1, \dots, n\}$. We write $\sM(\calX)$ to denote the output distribution of a randomized algorithm $\sM$. For a set $\mathcal{D}\subset\mathbb{R}^d$, we let $\textnormal{dia}(\mathcal{D})$ be its diameter, i.e., $\displaystyle \textnormal{dia}(\mathcal{D}) = \sup_{x_1,x_2\in\mathcal{D}} \lVert x_{2}-x_{1}\rVert$.
%%%%%%%

\section{Preliminaries}
\label{Section:Preliminaries}

% In this section, we recall relevant preliminaries about differential privacy and information theory. 

\subsection{Differential Privacy}

Differential privacy (DP) \cite{Dwork_Calibration} has become the de-facto standard for quantifying privacy in today's machine learning. Intuitively, DP measures how indistinguishable the outputs of a learning algorithm are when running on two \textit{neighboring} datasets, i.e., two datasets that differ in one entry. To quantify such indistinguishability, one can measure the distance between $\sM(\calX)$ and $\sM(\calX')$, output distributions of the mechanism $\sM$ when running on two neighboring datasets $\calX$ and $\calX'$.
The original definition of approximate differential privacy was based on the following ``distance'' measure, often called the \textit{hockey-stick divergence}.  
\begin{defn}[Hockey-stick divergence]\label{def:HSDivergence}
Given $\eps\geq 0$ and two probability measures $\mu$ and $\nu$ in $\calP(\calW)$, the hockey-stick divergence between $\mu$ and $\nu$ is defined as
$$\sE_\eps(\mu\|\nu) \coloneqq  \int \text{d}(\mu-e^\eps\nu)^+ = \sup_{A\subset \calW} ~[\mu(A) - e^\eps \nu(A)],$$%\frac{1}{2}\int|\text{d}\mu - e^\eps \text{d}\nu| - \frac{1}{2}(e^\eps - 1),$$
where $(\mu-e^\eps\nu)^+$ is the positive part of the signed measure $\mu-e^\eps\nu$.
\end{defn}
Observe that when $\eps =0$, the hockey-stick divergence reduces to total-variation distance, that is $\sE_0(\mu\|\nu) = \lVert \mu-\nu\rVert_{\mathrm{TV}}$.
With this definition at hand, we now define approximate differential privacy.
\begin{defn}[\cite{Dwork_Calibration}]
A randomized algorithm $\sM$ is said to be $(\eps, \delta)$-DP if, for any two neighboring datasets $\calX$ and $\calX'$, we have that $\sE_\eps(\sM(\calX)\Vert\sM(\calX'))\leq \delta$.
\end{defn}

Note that one can replace the hockey-stick divergence in the above definition with R\'enyi divergence to arrive at the definition of R\'enyi differential privacy (RDP) \cite{RenyiDP, ZeroDP}.

\subsection{Contraction Coefficient of Hockey-Stick Divergence}
\label{sec:ContractionCoeff}

It is known that the hockey-stick divergence satisfies the data processing inequality, that is, $\sE_\eps(\sK\mu\Vert\sK\nu)\leq \sE_\eps(\mu\Vert\nu)$ for any probability measures $\mu$ and $\nu$, and any Markov kernel $\sK$\footnote{It is worth noting that hockey-stick divergence is a special case of a larger family of divergences known as $f$-divergence, and that all $f$-divergences satisfy the data processing inequality. For more details, see \cite{Verdu:f_divergence, Raginsky_SDPI}}. For many non-trivial Markov kernels this inequality is strict. Consequently, one can think of a stronger data processing inequality by finding a constant $\eta \leq 1$ such that  $\sE_\eps(\sK\mu\Vert\sK\nu)\leq  \eta \sE_\eps(\mu\Vert\nu)$. The smallest such constant $\eta$ is called \textit{contraction coefficient} of hockey-stick divergence for Markov kernel $\sK$ and is denoted by $\eta_\eps(\sK)$. In other words, we have 
\begin{equation}\label{Eq:SDPI_HSDivergence}
    \eta_\eps(\sK)\coloneqq \sup_{\substack{\mu,\nu\in\calP(\calW):\\ \sE_\eps(\mu\|\nu)\neq 0}}\frac{\sE_\eps(\sK\mu\|\sK\nu)}{\sE_\eps(\mu\|\nu)}.
\end{equation}
If $\eta_\eps(\sK)<1$, then we say that $\sK$ satisfies a \textit{strong} data processing inequality (SDPI) for hockey-stick divergence with contraction coefficient $\eta_\eps(\sK)$. 

It was recently shown \cite[Theorem 2]{asoodeh2020online} that for a Markov kernel $\sK:\calW\to \calP(\mathbb{R}^d)$
$$\eta_\eps(\sK) = \sup_{x_1, x_2\in \calW}\sE_\eps(\sK(x_1)\Vert\sK(x_2)).$$ This result enables us to quantify privacy amplification by post-processing as follows: Given an $(\eps, \delta)$-DP mechanism $\sM$ and a Markov kernel $\sK$, we have for any neighboring datasets $\calX$ and $\calX'$
$$\sE_\eps(\sK\sM(\calX)\Vert\sK\sM(\calX'))\leq \sE_\eps(\sM(\calX)\Vert\sM(\calX'))  \eta_\eps(\sK) \leq \delta  \sup_{x_1, x_2\in \calW}\sE_\eps(\sK(x_1)\Vert\sK(x_2)),$$
indicating $\sM\circ \sK$ is $(\eps, \delta')$-DP with $\delta' = \delta \sup_{x_1, x_2\in \calW}\sE_\eps(\sK(x_1)\Vert\sK(x_2))$. Thus, characterizing privacy amplification by post-processing boils down to computing the contraction coefficient of the post-processor. 
If post-processor $\sK$ is a Gaussian Markov kernel with bounded input, then $\eta_\eps(\sK)$ enjoys a simple closed-form expression, as delineated by the following proposition.  Given $\calW\subset\mathbb{R}^{d}$ and $\sigma>0$, we define 
the \textit{$\calW$-constrained Gaussian kernel} $\sK_{\mathrm{G}}^\sigma:\calW\to\calP(\mathbb{R}^d)$ as
\begin{equation*}
    \sK_{\mathrm{G}}^\sigma(x) = \mathcal{N}(x,\sigma^2 \mathbf{I}_d).
\end{equation*}
\begin{proposition}[\cite{asoodeh2020online}]
\label{Proposition:ContractionCoefficientGaussianKernel}
Let $\calW\subset\mathbb{R}^{d}$ and $\sigma\geq0$. For any $\calW$-constrained Gaussian kernel $\sK_{\mathrm{G}}^\sigma$, we have
\begin{equation*}
    \eta_{\eps}(\sK_{\mathrm{G}}^\sigma) = \theta_{\eps}\Big(\frac{\mathrm{dia}(\calW)}{\sigma}\Big),
\end{equation*}
where $$\theta_{\eps}(r) \coloneqq Q\Big(\frac{\eps}{r} - \frac{r}{2}\Big) - e^\eps Q\Big(\frac{\eps}{r} + \frac{r}{2}\Big),$$ 
with $\displaystyle Q(t) \coloneqq \frac{1}{\sqrt{2\pi}} \int_{t}^{\infty} e^{-u^{2}/2} \mathrm{d}u$.
\end{proposition}
In light of this result, $\calW$-constrained Gaussian kernels satisfy SDPI  for hockey-stick divergence if $\text{dia}(\calW)<\infty$.   We will repeatedly use this observation in the next section to derive a diameter-aware bound on the privacy parameters of \textsf{DP-SGD}.
%%%%%%%

\section{Main Results}
\label{Section:MainResults}

In this section, we begin with a general iterative process which  we later instantiate for different versions of \textsf{DP-SGD}.
Let $\sigma\geq0$ and $\mathcal{W}_{t}\subseteq\mathbb{R}^{d}$ for each $t \leq T$. We consider the iterative process $\{W_{t}\}_{t \leq T}$ determined by $W_{0} \sim \mu_{0}\in\mathcal{P}(\mathcal{W}_{0})$ and
\begin{equation}
\label{eq:GeneralUpdateRule}
    W_{t} = \Pi_{t}\big[\psi_{B_{t}}(W_{t-1}) + \sigma Z_{t}\big],
\end{equation}
where $\Pi_{t}:\mathbb{R}^{d}\to\mathcal{W}_{t}$ is the projection onto $\mathcal{W}_{t}$, $Z_{t}$ is a standard Gaussian vector independent of all other variables, $B_{t}$ is a random batch, and $\psi_{B}:\mathbb{R}^{d}\to\mathbb{R}^{d}$ is an update function dependent on the dataset $\mathcal{X}$ through the batch $B\subset[n]$. In particular, the update kernel $\sK_{t} \coloneqq P_{W_{t} \vert W_{t-1}}$ could be decomposed as
\begin{equation}
\label{eq:DefinitionKt}
    \sK_{t} = \Pi_{t} \circ \sK_{\mathrm{G}}^{\sigma} \circ \Psi,
\end{equation}
where $\Psi:\mathbb{R}^{d}\to\mathcal{P}(\mathbb{R}^{d})$ is the kernel given by $\Psi = \sum_{B\subset[n]} \mathbb{P}(B_{t} = B) \psi_{B}$. Recall that, by abuse of notation, a deterministic function $\psi_B$ can be viewed as a Markov kernel determined by $w\mapsto \delta_{\psi_B(w)}$.

Observe that the iterative process \eqref{eq:GeneralUpdateRule} is $(\eps, \delta)$-differentially private after $T$ iterations if
\begin{equation}
\label{eq:MainResultsHSDivergenceT}
    \sE_\eps(\sK_T\cdots\sK_1\mu_0\|\sK'_T\cdots\sK'_1\mu_0)\leq \delta,
\end{equation}
where $\sK_1, \dots, \sK_T$ and $\sK'_1, \dots, \sK'_T$ are the update kernels in \eqref{eq:DefinitionKt} corresponding to neighboring datasets $\calX$ and $\calX'$, respectively. The following result presents a machinery to bound the hockey-stick divergence in \eqref{eq:MainResultsHSDivergenceT} using  bounds for individual update kernels $\sK_{t}$.

\begin{lemma}
\label{Lemma:HSDivergenceRecursion}
Let $\mathcal{X}$ and $\mathcal{X}'$ be two neighboring datasets. Let $\{\sK_{t}\}_{t\in[T]}$ and $\{\sK_{t}'\}_{t\in[T]}$ be defined as in \eqref{eq:DefinitionKt} for $\mathcal{X}$ and $\mathcal{X}'$, respectively. If for each $t\in[T]$ there exist $\alpha_{t},\beta_{t}\in[0,1]$ such that
\begin{equation}
\label{eq:CoupledNonlinearSDPI}
    \phantom{\forall \mu,\nu\in\mathcal{P}(\mathcal{W_{t-1}}),} \qquad \sE_\eps(\sK_{t} \mu \Vert \sK_{t}' \nu) \leq \alpha_{t} \sE_\eps(\mu \Vert \nu) + \beta_{t}, \qquad \forall \mu,\nu\in\mathcal{P}(\mathcal{W}_{t-1}),
\end{equation}
then
\begin{equation}
\label{eq:LemmaHSDivergenceRecursion}
    \sE_\eps\big(\sK_{T} \cdots \sK_{1} \mu \Vert \sK_{T}' \cdots \sK_{1}' \nu\big) \leq \Big[\prod_{t=1}^{T} \alpha_{t}\Big] \sE_\eps(\mu \Vert \nu) + \sum_{t=1}^{T} \beta_{t} \prod_{s=t+1}^{T} \alpha_{s}.
\end{equation}
\end{lemma}

The proof of this lemma in Appendix~\ref{Appendix:ProofsMainResults} exploits the recursive nature of \eqref{eq:CoupledNonlinearSDPI}. We remark that Lemma~\ref{Lemma:HSDivergenceRecursion} resembles    \cite[Lemma~3.2]{Shokri_SGD_SGLD}, in that both present recursive relations for computing the divergence between output distributions of process \eqref{eq:GeneralUpdateRule} when running on two neighboring datasets. 
%that the idea of studying divergences in a recursive manner has been used in the literature before. For example, Ye and Shokri studied the R\'{e}nyi divergence associated with noisy SGD in \cite[Lemma~3.2]{Shokri_SGD_SGLD}.

It is also worth noting that Lemma~\ref{Lemma:HSDivergenceRecursion} differs from the framework based on the strong data processing inequality (SDPI) proposed by Asoodeh \textit{et al.} in \cite{asoodeh2020online}. First, the kernels $\sK_{t}$ and $\sK_{t}'$ in \eqref{eq:CoupledNonlinearSDPI} are potentially different (although they could share some common randomness), and thus $\sE_\eps(\sK_t\mu\|\sK_t'\nu)$ cannot be bounded directly via contraction coefficients. Second, the existence of the additive term $\beta_{t}$ in \eqref{eq:CoupledNonlinearSDPI} renders the upper bound in \eqref{eq:LemmaHSDivergenceRecursion} fundamentally different from the multiplicative bounds in \cite[Theorem 1]{asoodeh2020online}. In this sense, condition \eqref{eq:CoupledNonlinearSDPI} could be regarded as a \textit{coupled non-linear DPI}. In the sequel, we prove that such condition holds for different versions of \textsf{DP-SGD}, leading to the convergence of the associated privacy losses.
%%%

\subsection{Projected DP-SGD}

In this section, we consider a projected version of the algorithm known as \textsf{DP-SGD} introduced by Abadi \textit{et al.} in \cite{abadi2016deep}. As it is customary, we take $\mathcal{W}_{t} = \mathcal{W}$ for some fixed convex set $\mathcal{W}\subset\mathbb{R}^{d}$ and let $\Pi$ be the projection onto $\mathcal{W}$. For a dataset $\mathcal{X} = \{x_{1},\ldots,x_{n}\}$, we define the update mapping with respect to a batch $B\subset[n]$ as
\begin{equation}
\label{eq:DefPsiBProjected}
    \psi_{B}(w) = w - \frac{\eta}{\lvert B \rvert} \sum_{i\in B} \mathrm{Clip}\big(\nabla\ell(w,x_{i})\big).
\end{equation}
where $\mathrm{Clip}:\mathbb{R}^{d}\to\mathbb{R}^{d}$ is the clipping function, defined in \eqref{eq:DefClipping}, with clipping constant $C$.

The following lemma, whose proof could be found in Appendix~\ref{Appendix:ProofsMainResults}, establishes a key boundedness property of projected \textsf{DP-SGD}.

\begin{lemma}
\label{Lemma:ProjectedDPSGDDomain}
Under projected \textsf{DP-SGD}, we have that
\begin{equation*}
    \sup_{B_{1},B_{2}\subset[n]} \sup_{w_{1},w_{2}\in\mathcal{W}} \lVert \psi_{B_{2}}(w_{2}) - \psi_{B_{1}}(w_{1}) \rVert \leq D+2 \eta C,
\end{equation*}
where $D = \mathrm{dia}(\mathcal{W})$, $\eta$ is the learning rate, and $C$ is the clipping constant.
\end{lemma}

In the sequel, we prove that the coupled non-linear DPI in \eqref{eq:CoupledNonlinearSDPI} is satisfied for projected \textsf{DP-SGD} under both Poisson sampling and sampling without replacement.

\subsubsection{Poisson Sampling}

In this sampling scheme, each datapoint is selected with probability $p\in[0,1]$ at each iteration. Thus, the average batch size is $b = pn$. Under Poisson sampling, the kernel $\Psi$ in \eqref{eq:DefinitionKt} is given by
\begin{equation}
\label{eq:DefPsiPoissonSampling}
    \Psi = \sum_{B\subset[n]} p^{\lvert B \rvert} (1-p)^{n-\lvert B \rvert} \psi_{B}.
\end{equation}
The following proposition establishes the coupled non-linear DPI in \eqref{eq:CoupledNonlinearSDPI} for the kernel associated with projected \textsf{DP-SGD} under Poisson sampling.

\begin{proposition}
\label{Proposition:CoupledNonlinearDPIProjectedPoisson}
Let $\mathcal{X}$ and $\mathcal{X}'$ be two neighboring datasets. Let $\Psi$ and $\Psi'$ be defined as in \eqref{eq:DefPsiPoissonSampling} for $\mathcal{X}$ and $\mathcal{X}'$, respectively. If $\sK = \Pi \circ \sK_{\mathrm{G}}^{\sigma} \circ \Psi$ and $\sK' = \Pi \circ \sK_{\mathrm{G}}^{\sigma} \circ \Psi'$, then, for all $\mu,\nu\in\mathcal{P}(\mathcal{W})$,
\begin{equation}
\label{eq:CoupledNonlinearDPIProjectedPoisson}
    \sE_\eps(\sK\mu \Vert \sK'\nu) \leq (1-p) \theta_\eps\bigg(\frac{D+2\eta C}{\sigma}\bigg) E_\eps(\mu \Vert \nu) + p\theta_\eps\bigg(\frac{D+2\eta C}{\sigma}\bigg).
\end{equation}
\end{proposition}

The proof of this proposition in Appendix~\ref{Appendix:ProofsMainResults} indicates that the multiplicative term in \eqref{eq:CoupledNonlinearDPIProjectedPoisson} comes from the batches that do not contain the different point in $\mathcal{X}$ and $\mathcal{X}'$ and the additive term comes from the batches that contain such a point. In both cases, the strong data processing inequality for input-constrained Gaussian kernels in Proposition~\ref{Proposition:ContractionCoefficientGaussianKernel} produces the factor $\theta_{\eps}$.

The previous proposition and Lemma~\ref{Lemma:HSDivergenceRecursion} enable us to characterize the privacy guarantees of \textsf{DP-SGD} under Poisson sampling.

\begin{theorem}\label{Thm:DP_SGD_Poisson}
Let $\eps\geq0$. After $T\in\mathbb{N}$ iterations, projected \textsf{DP-SGD} under Poisson sampling is $(\eps,\delta)$-differentially private with
\begin{equation*}
    \delta \leq \frac{1-[(1-p)\theta]^{T}}{1-(1-p)\theta} p\theta,
\end{equation*}
where $\theta \coloneqq \theta_\eps\big((D+2\eta C)/\sigma\big)$.
\end{theorem}

\subsubsection{Sampling without Replacement}

In this sampling scheme,  a random sample of size $b \leq n$ is selected at each iteration. Thus, any batch $B\subseteq[n]$ of size $b$ is selected with probability $\binom{n}{b}^{-1}$. Under the sampling without replacement, the kernel $\Psi$ in \eqref{eq:DefinitionKt} is given by
\begin{equation}
\label{eq:DefPsiSampleWithoutReplacement}
    \Psi = \frac{1}{\binom{n}{b}} \sum_{B\subset[n]: \lvert B \rvert = b} \psi_{B}.
\end{equation}
The following proposition establishes the coupled non-linear DPI in \eqref{eq:CoupledNonlinearSDPI} for the kernel associated with projected \textsf{DP-SGD} under sampling without replacement.

\begin{proposition}
\label{Proposition:CoupledNonlinearDPIProjectedSWR}
Let $\mathcal{X}$ and $\mathcal{X}'$ be two neighboring datasets. Let $\Psi$ and $\Psi'$ be defined as in \eqref{eq:DefPsiSampleWithoutReplacement} for $\mathcal{X}$ and $\mathcal{X}'$, respectively. If $\sK = \Pi \circ \sK_{\mathrm{G}}^{\sigma} \circ \Psi$ and $\sK' = \Pi \circ \sK_{\mathrm{G}}^{\sigma} \circ \Psi'$, then, for all $\mu,\nu\in\mathcal{P}(\mathcal{W})$,
\begin{equation*}
    \sE_\eps(\sK \mu \Vert \sK' \nu) \leq \bigg(1-\frac{b}{n}\bigg) \theta_\eps\bigg(\frac{D+2\eta C}{\sigma}\bigg) \sE_\eps(\mu \Vert \nu) + \frac{b}{n}\theta_\eps\bigg(\frac{D+2\eta C}{\sigma}\bigg).
\end{equation*}
\end{proposition}

The proof of the previous proposition, which can be found in Appendix~\ref{Appendix:ProofsMainResults}, is similar to the proof of Proposition~\ref{Proposition:CoupledNonlinearDPIProjectedPoisson}. Together with Lemma~\ref{Lemma:HSDivergenceRecursion}, Proposition~\ref{Proposition:CoupledNonlinearDPIProjectedSWR} enable us to characterize the privacy guarantees of \textsf{DP-SGD} under sampling without replacement.

\begin{theorem}\label{Thm:DP_SGD_Replacement}
Let $\eps\geq0$. After $T\in\mathbb{N}$ iterations, the projected \textsf{DP-SGD} under sampling without replacement is $(\eps,\delta)$-differentially private with
\begin{equation*}
    \delta \leq \frac{1-[(1-p)\theta]^{T}}{1-(1-p)\theta} p\theta,
\end{equation*}
where $p \coloneqq b/n$ and $\theta \coloneqq \theta_\eps\big((D+2\eta C)/\sigma\big)$.
\end{theorem}

Comparing this theorem with Theorem~\ref{Thm:DP_SGD_Poisson}, we observe that \textsf{DP-SGD} under sampling without replacement with batch size $b$ has the same privacy guarantees as \textsf{DP-SGD} under Poisson sampling with sub-sampling probability $b/n$.

Theorems~\ref{Thm:DP_SGD_Poisson} and \ref{Thm:DP_SGD_Replacement} identify the privacy guarantees of projected \textsf{DP-SGD}, where the update at each iteration is projected onto a given parameter space $\calW$. In the next section, we seek to study the privacy guarantees of \textit{regularized} (unprojected) \textsf{DP-SGD} using the framework established in Lemma~\ref{Lemma:HSDivergenceRecursion}.
%%%

\subsection{Regularized DP-SGD}

In this section, we analyze the privacy loss of regularized \textsf{DP-SGD}. For ease of exposition, in the sequel we only consider sampling without replacement. The analysis of Poisson sampling is similar, so the details are left to the reader.

Let $\lambda\in(0,1)$. We consider the iterative process $\{\tilde{W}_{t}\}_{t \leq T}$ determined by\footnote{The assumption $\tilde{W}_{0}=0$ is immaterial but we adopted it for ease of exposition.} $\tilde{W}_{0} = 0$ and
\begin{equation}
\label{eq:GeneralUpdateRuleRegularized}
    \tilde{W}_{t} = \psi_{B_{t}}(\tilde{W}_{t-1}) + \sigma Z_{t},
\end{equation}
where $B_{t}$ is a random batch of size $b$ and, for each $B\subset[n]$,
\begin{equation}
\label{eq:DefPsiBRegularized}
    \psi_{B}(w) = (1-\lambda)w - \frac{\eta}{\lvert B \rvert} \sum_{i\in B} \mathrm{Clip}\big(\nabla\ell(w,x_{i})\big).
\end{equation}
Note that we assume that the regularization term (i.e., $\lambda w$) is added after clipping, as considered by Ye and Shokri in their recent work \cite{Shokri_SGD_SGLD}. Since there is no projection in this case, the update kernel $\tilde{\sK}_{t} \coloneqq P_{\tilde{W}_{t} \vert \tilde{W}_{t-1}}$ could be decomposed as
\begin{equation}
\label{eq:RegularizedDefPsi}
    \tilde{\sK}_{t} = \sK_{\mathrm{G}}^{\sigma} \circ \Psi \qquad \text{where} \qquad \Psi = \frac{1}{\binom{n}{b}} \sum_{B\subset[n]: \lvert B \rvert = b} \psi_{B}.
\end{equation}
We study the privacy loss of \eqref{eq:GeneralUpdateRuleRegularized} by relying on the particular instance of the iterative process  \eqref{eq:GeneralUpdateRule} described next.

Let $\kappa>0$. We consider the iterative process $\{W_{t}\}_{t \leq T}$ determined by $W_{0} = 0$ and
\begin{equation}
\label{eq:GeneralUpdateRuleRegularizedProjected}
    W_{t} = \Pi_{t}\big[\psi_{B_{t}}(W_{t-1}) + \sigma Z_{t}\big],
\end{equation}
where $\Pi_{t}:\mathbb{R}^{d}\to\mathcal{W}_{t}$ is the projection onto $\mathcal{W}_{t} \coloneqq \mathbb{B}_{d}(r_{t})= \{w\in \mathbb{R}^d: \|w\|\leq r_t\}$ with
\begin{equation}
\label{eq:RegularizedRadious}
    r_{t} = (\eta C + \kappa \sigma)\frac{1-(1-\lambda)^{t}}{\lambda}.
\end{equation}
Note that the update kernel $\sK_{t} \coloneqq P_{W_{t} \vert W_{t-1}}$ could be decomposed as $\sK_{t} = \Pi_{t} \circ \sK_{\mathrm{G}}^{\sigma} \circ \Psi$.
The introduction of the iterative process $\{W_{t}\}_{t \leq T}$ is motivated by the following three intuitions:
\begin{itemize}
    \item[(i)] Since $\sK_{t}$ and $\tilde{\sK}_{t}$ differ only by the projection step $\Pi_{t}$, by making $\kappa$ large enough this projection becomes immaterial, i.e., if $\kappa$ is large enough then $\lVert W_{t} \rVert \leq r_{t}$ with high probability. In this case, the distributions of $\{W_{t}\}_{t \leq T}$ and $\{\tilde{W}_{t}\}_{t \leq T}$ are close.

    \item[(ii)] The projection step $\Pi_{t}$ in $\sK_{t}$ allows us to establish a coupled non-linear DPI similar to the one derived for projected \textsf{DP-SGD} in Proposition~\ref{Proposition:CoupledNonlinearDPIProjectedSWR}. This, together with Lemma~\ref{Lemma:HSDivergenceRecursion}, enables us to characterize the privacy guarantees of $\{W_{t}\}_{t \leq T}$.

    \item[(iii)] Upon a triangle inequality for the hockey-stick divergence by Liu \textit{et al.}\ \cite{liu2016e}, we can transfer the privacy guarantees of $\{W_{t}\}_{t \leq T}$ to $\{\tilde{W}_{t}\}_{t \leq T}$.
\end{itemize}
The following proposition formalizes the first intuition.
\begin{proposition}
\label{Prop:RegularizedCouplingControl}
Let $\chi^{2}_{d}$ be a chi-squared random variable with $d$ degrees of freedom. If $W_{t}\sim \mu_{t}$ and $\tilde{W}_{t} \sim \tilde{\mu}_{t}$, then $\displaystyle \lVert \tilde{\mu}_{t} - \mu_{t} \rVert_{\mathrm{TV}} \leq 1-\mathbb{P}(\chi^{2}_{d} \leq \kappa^{2})^{t}$ for all $t\geq0$.
\end{proposition}

As anticipated, sufficiently large $\kappa$ implies that the distributions of $\{W_{t}\}_{t \leq T}$ and $\{\tilde{W}_{t}\}_{t \leq T}$ are close. The next proposition provides a coupled non-linear DPI for $\{W_{t}\}_{t \leq T}$, formalizing the second intuition above.

\begin{proposition}
\label{Proposition:RegularizedCoupledNonLinearSDPI}
Let $\mathcal{X}$ and $\mathcal{X}'$ be two neighboring datasets. For each $B\subset[n]$, let $\psi_{B}$ and $\psi_{B}'$ be defined as in \eqref{eq:DefPsiBRegularized} for $\mathcal{X}$ and $\mathcal{X}'$, respectively. If $\sK_t = \Pi_{t} \circ \sK_{\mathrm{G}}^{\sigma} \circ \Psi$ and $\sK_t' = \Pi_{t} \circ \sK_{\mathrm{G}}^{\sigma} \circ \Psi'$ where $\Pi_{t}$ is the projection onto $\calW_t=\mathbb{B}_{d}(r_{t})$, then, for all $\mu,\nu\in\mathcal{P}(\mathcal{W}_{t-1})$,
\begin{equation*}
    \sE_\eps(\sK_t\mu \Vert \sK_t'\nu) \leq \bigg(1-\frac{b}{n}\bigg) \theta_\eps\bigg(\frac{r_{t}-\sigma\kappa}{\sigma}\bigg) \sE_\eps(\mu \Vert \nu) + \frac{b}{n} \theta_\eps\bigg(\frac{r_{t}-\sigma\kappa}{\sigma}\bigg).
\end{equation*}
\end{proposition}

Finally, we obtain the privacy loss of $\{\tilde{W}_{t}\}_{t \leq T}$ upon Propositions~\ref{Prop:RegularizedCouplingControl} and \ref{Proposition:RegularizedCoupledNonLinearSDPI}. The proof of the following theorem, which can be found in Appendix~\ref{Appendix:ProofsMainResults}, relies on a triangle inequality for the hockey-stick divergence proved by Liu \textit{et al.}\ in \cite{liu2016e}.

\begin{theorem}
\label{Theorem:RegularizedDPSGD}
Let $\eps\geq0$ and $\kappa>0$. After $T\in\mathbb{N}$ iterations, regularized \textsf{DP-SGD} under sampling without replacement is $(\eps,\delta)$-differentially private with
\begin{equation}
\label{eq:RegularizedDPSGD}
    \delta \leq (1+e^{\eps})\Big[1-\mathbb{P}(\chi^{2}_{d} \leq \kappa^{2})^{T}\Big] + \frac{b}{n-b} \sum_{t=1}^{T} \bigg(1-\frac{b}{n}\bigg)^{T-t} \prod_{s=t}^{T} \theta_\eps\bigg(\frac{r_{s} - \sigma\kappa}{\sigma}\bigg),
\end{equation}
where $\displaystyle r_{s} = (\eta C + \kappa \sigma)\frac{1-(1-\lambda)^{s}}{\lambda}$.
\end{theorem}

\begin{figure}[t]
   \hspace{-15pt}
   \begin{subfigure}{.4\textwidth}
    \includegraphics[scale=.4]{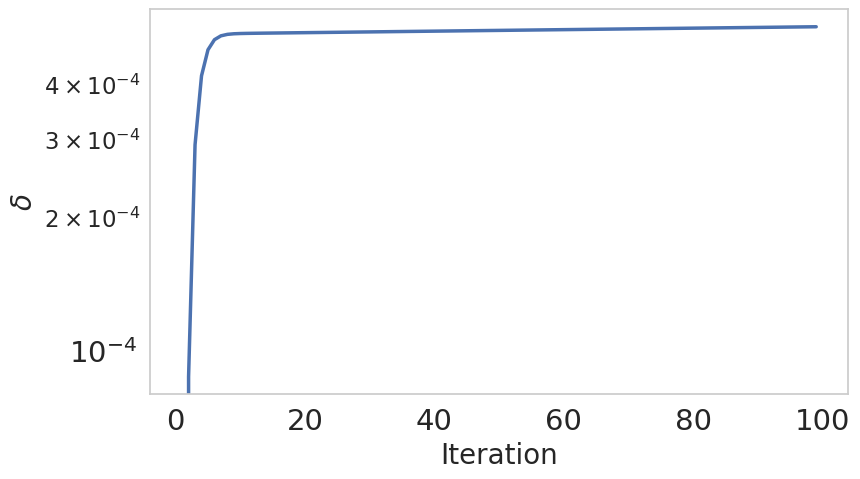}
    \end{subfigure}
    \qquad \qquad\quad
     \begin{subfigure}{.4\textwidth}
    \includegraphics[scale=.4]{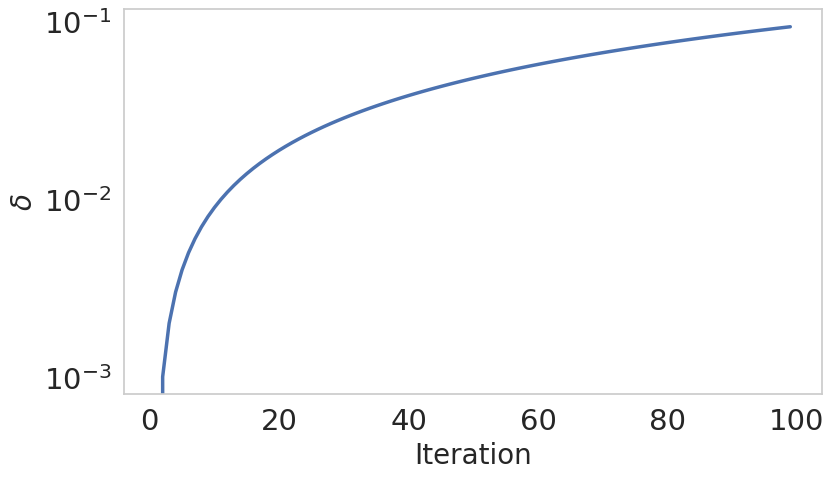}
    \end{subfigure}
    \caption{The privacy loss of regularized unprojected \textsf{DP-SGD} according to Theorem~\ref{Theorem:RegularizedDPSGD} with parameters: $C = 1, \eta = 0.1, \sigma = 5, \eps = 5,p = 0.001$, and $\lambda = 0.65$ (left) and $\lambda=0.01$ (right).  }
    \label{fig:regularized}
\end{figure}

Note that \eqref{eq:RegularizedDPSGD} holds for every $\kappa>0$. Hence, this parameter has to be optimized (numerically) in order to obtain the tightest bound. However, the numerical optimization of \eqref{eq:RegularizedDPSGD} could be subtle for some range of parameters. Therefore, it is unclear if the optimized bound in Theorem~\ref{Theorem:RegularizedDPSGD} converges with respect to the number of iterations. In fact, Figure~\ref{fig:regularized} shows that even numerically such convergence is unclear. Nonetheless, for some parameter values the bound produces reasonable estimates for the privacy loss.

%%%%%%%

\section{Discussion}
\label{Section:Discussion}
The framework developed in this work assumes that the Gaussian Markov kernel that model each iteration of \textsf{Noisy-SGD} have bounded input. This can be realized by either gradient clipping or by running \textsf{Noisy-SGD} on the smoothed version of the objective (e.g., by Gaussian convolution or Moreau-Yousida smoothing). This enables us to exploit the probabilistic contraction of Gaussian kernels to prove convergence of the privacy loss under minimal assumptions on losses. %The following remarks about our technique are noteworthy:   
%\begin{itemize}
    %\item
    
    It is worth mentioning that typical regularity assumptions on the loss function (such as convexity and smoothness) would improve our results. For instance, it can be shown via geometric contraction arguments in  \cite{Feldman2018PrivacyAB,Balle2019mixing} that the map $w\mapsto w - \eta \nabla\ell(w, x)$ is contractive for smooth and (strictly) convex loss functions. Consequently, the diameter of the input of Gaussian kernel in \textsf{Noisy-SGD} is (strictly) smaller than $\text{dia}(\calW)$, thus leading to tighter bounds on the contraction coefficients of Gaussian kernels. 
    %\item
    
    As mentioned earlier, it is not clear that our technique leads to a convergent upper bound for the regularized \textsf{DP-SGD}. A natural future direction is to improve Theorem~\ref{Theorem:RegularizedDPSGD} (perhaps by means of a different coupling process) and analytically prove the convergence of privacy loss in this setting.  
%\end{itemize}

%Geometric Contraction vs Probabilistic Contraction and what we learned from the regularized case.

%Regularized not clear, numerically or theoretically.

%Advantages and limitations of using HS-divergence directly (no conversion, almost no restrictions on the parameter space, ...) (very worst case)

% Our results under more assumption: we are weaker but more general. It is worth noting that if we make more assumptions about the loss function, then the 
% resulting bound on $\eta_\eps(\sK)$ will be tighter. For instance, if the loss function is additionally convex, then it is known that $w_t\mapsto w_t - \eta G_t$ is $1$-Lipschitz \cite{Nesterov:ILC} for $\eta\leq \frac{2}{M}$ and thus the diameter of the input domain of Gaussian mechanism in $\sK_t$ is $D$ instead of $(1 + \frac{Mb\eta}{n})D$.  If instead, the loss function is $\rho$-strongly convex, then $w_t\mapsto w_t - \eta G_t$ is $L$-Lipschitz with $L = \sqrt{1-\frac{2M\eta\rho}{M+\rho}}<1$ for $\eta\leq \frac{2}{M+\rho}$ \cite[Lemma 17]{Balle2019mixing}. Thus, the input of Gaussian mechanism in each iteration has a diameter strictly smaller than $D$.
%%%%%%%

%\begin{comment}

\appendix
\section{Conversion of Privacy Loss Bounds}
\label{Appendix:ConversionPrivacyLossBounds}

Recall that according to Main Theorem (or Theorem~\ref{Thm:DP_SGD_Replacement}), the projected \textsf{DP-SGD} under sampling without replacement with batch size $b$ is $(\eps,\delta)$-differentially private with $\eps\geq 0$ and
\begin{equation*}
    \delta \leq \frac{1-[(1-p)\theta]^{T}}{1-(1-p)\theta} p\theta,
\end{equation*}
where $p \coloneqq \frac{b}{n}$, 
$\theta \coloneqq \theta_\eps\big(\frac{D+2\eta C}{\sigma}\big)$ and 
$$\theta_{\eps}(r) \coloneqq Q\Big(\frac{\eps}{r} - \frac{r}{2}\Big) - e^\eps Q\Big(\frac{\eps}{r} + \frac{r}{2}\Big)$$
with $\displaystyle Q(t) \coloneqq \frac{1}{\sqrt{2\pi}} \int_{t}^{\infty} e^{-u^{2}/2} \mathrm{d}u$. 

It can be easily verified that the above upper bound for $\delta$ monotonically converges to $\displaystyle \frac{p\theta}{1-(1-p)\theta}$ as $T$ goes to infinity. Thus,  it follows that 
\begin{equation}\label{eq:eps_UB1}
    \eps \leq \eps^*_\delta, 
\end{equation}
where $\eps^*_\delta$ is the unique $\eps$ that solves $$ \theta_{\eps}(r) = \frac{\delta}{p + (1-p)\delta},$$ 
with $r \coloneqq \frac{D+2\eta C}{\sigma}$. Unfortunately,  $\eps^*_\delta$ cannot be computed in closed-form; however, we can obtain the following upper bound by noticing that $\theta_\eps(r) < Q\big(\frac{\eps}{r} - \frac{r}{2}\big)$:
\begin{equation}\label{eq:eps_UB2}
    \eps\leq r\left[\frac{r}{2} + \Phi^{-1}\bigg(\frac{p(1-\delta)}{p + (1-p)\delta}\bigg)\right],
\end{equation}
where $\Phi^{-1}$ is the inverse of CDF of standard Gaussian distribution. 
\begin{figure}
    \centering
    \includegraphics[scale=0.45]{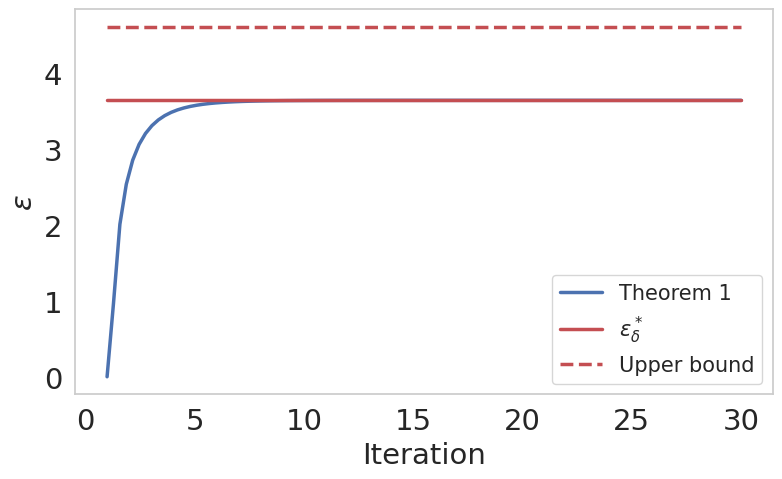}
    \caption{The privacy loss of \textsf{DP-SGD} as described in Theorem~\ref{Thm:DP_SGD_Replacement}  with the following parameters: $D = 3, C = 2, \eta = 0.01, \sigma = 1, \delta = 10^{-3}$ and $p = 0.001$. Red curves correspond to the upper bounds on $\eps$ derived in \eqref{eq:eps_UB1} and \eqref{eq:eps_UB2}.}
    \label{fig:my_label}
\end{figure}
%%%%%%%

\section{Proofs of the Main Results}
\label{Appendix:ProofsMainResults}

\begin{proof}[\textbf{Proof of Lemma~\ref{Lemma:HSDivergenceRecursion}}]
Define $\mu_{T-1} \coloneqq \sK_{T-1} \cdots \sK_{1} \mu$ and $\nu_{T-1} = \sK'_{T-1} \cdots \sK'_{1} \nu$. Due to the projection $\Pi_{T-1}$ present in both $\sK_{T-1}$ and $\sK_{T-1}'$, we have that $\mu_{T-1}, \nu_{T-1}\in \calP(\calW_{T-1})$. Hence, \eqref{eq:CoupledNonlinearSDPI} implies that
\begin{align*}
    \sE_\eps\big(\sK_{T} \cdots \sK_{1} \mu \Vert \sK_{T}' \cdots \sK_{1}' \nu\big) & = \sE_\eps\big(\sK_{T}\mu_{T-1} \Vert \sK_{T}'\nu_{T-1}\big)\nonumber\\
    &\leq \alpha_T\sE_\eps(\mu_{T-1} \Vert \nu_{T-1}) + \beta_{T}.%\label{Proof1_lemmaRecursive}
\end{align*}
By applying the same argument iteratively, we obtain that
\begin{align*}
    \sE_\eps\big(\sK_{T} \cdots \sK_{1} \mu \Vert \sK_{T}' \cdots \sK_{1}' \nu\big) &\leq \alpha_{T} \sE_\eps\big(\sK_{T-1} \cdots \sK_{1} \mu \Vert \sK_{T-1}' \cdots \sK_{1}' \nu\big) + \beta_{T}\\
    &\leq \alpha_{T-1}\alpha_{T} \sE_\eps\big(\sK_{T-2} \cdots \sK_{1} \mu \Vert \sK_{T-2}' \cdots \sK_{1}' \nu\big) + \beta_{T-1} \alpha_{T} + \beta_{T}\\
    &\;\;\vdots\\
    &\leq \Big[\prod_{t=1}^{T} \alpha_{t}\Big] \sE_\eps(\mu \Vert \nu) + \sum_{t=1}^{T} \beta_{t} \prod_{s=t+1}^{T} \alpha_{s},
\end{align*}
as desired.
\end{proof}

\begin{proof}[\textbf{Proof of Lemma~\ref{Lemma:ProjectedDPSGDDomain}}]
From the definition of $\mathrm{Clip}:\mathbb{R}^{d}\to\mathbb{R}^{d}$ in \eqref{eq:DefClipping}, we have that $\lVert \mathrm{Clip}(v) \rVert \leq C$ for any $v\in\mathbb{R}^d$. Hence, \eqref{eq:DefPsiBProjected} implies that, for all $B\subset[n]$ and $w\in\mathcal{W}$,
\begin{equation*}
    \lVert \psi_{B}(w) - w \rVert \leq \eta C.
\end{equation*}
Hence, the triangle inequality implies that, for all $B_{1},B_{2}\subset[n]$ and $w_{1},w_{2}\in\mathcal{W}$,
\begin{align*}
    \lVert \psi_{B_{2}}(w_{2}) - \psi_{B_{1}}(w_{1}) \rVert &\leq \lVert \psi_{B_{2}}(w_{2}) - w_{2} \rVert + \lVert w_{2} - w_{1} \rVert + \lVert w_{1} - \psi_{B_{1}}(w_{1}) \rVert\\
    &\leq \mathrm{dia}(\mathcal{W})+2 \eta C.
\end{align*}
Since $B_{1},B_{2}\subset[n]$ and $w_{1},w_{2}\in\mathcal{W}$ were arbitrary, the result follows.
\end{proof}

\begin{proof}[\textbf{Proof of Proposition~\ref{Proposition:CoupledNonlinearDPIProjectedPoisson}}]
Let $\mathcal{X} = \{x_{1},\ldots,x_{n}\}$ and $\mathcal{X}' = \{x'_{1},\ldots,x'_{n}\}$. Without loss of generality, we may assume that $x_{n} \neq x'_{n}$ and $x_i = x'_i$ for $i\in [n-1]$. Recall the definition of $\Psi$ in \eqref{eq:DefPsiPoissonSampling}. In particular, we have that
\begin{equation*}
    \sK = \sum_{B\subset[n]} p^{\lvert B \rvert} (1-p)^{n-\lvert B \rvert} \Pi \circ \sK^\sigma_{\mathrm{G}} \circ \psi_{B} \qquad \text{and} \qquad \sK' = \sum_{B\subset[n]} p^{\lvert B \rvert} (1-p)^{n-\lvert B \rvert} \Pi \circ \sK^\sigma_{\mathrm{G}} \circ \psi_{B}'.
\end{equation*}
By the convexity of $(P,Q) \mapsto \sE_\eps(P \Vert Q)$, the previous expressions lead to
\begin{align*}
    \sE_\eps(\sK\mu \Vert \sK'\nu) &\leq \sum_{B\subset[n]} p^{\lvert B \rvert} (1-p)^{n-\lvert B \rvert} \sE_\eps\big((\Pi \circ \sK_{\mathrm{G}}^{\sigma} \circ \psi_{B}) \mu \Vert (\Pi \circ \sK_{\mathrm{G}}^{\sigma} \circ \psi_{B}') \nu\big)\\
    &= \sum_{B\subset[n]} p^{\lvert B \rvert} (1-p)^{n-\lvert B \rvert} \sE_\eps\big((\Pi \circ \sK_{\mathrm{G}}^{\sigma})(\psi_{B}\mu) \Vert (\Pi \circ \sK_{\mathrm{G}}^{\sigma})(\psi_{B}'\nu)\big)\\
    &\leq \sum_{B\subset[n]} p^{\lvert B \rvert} (1-p)^{n-\lvert B \rvert} \sE_\eps\big(\sK_{\mathrm{G}}^{\sigma}(\psi_{B}\mu) \Vert \sK_{\mathrm{G}}^{\sigma}(\psi_{B}'\nu)\big),
\end{align*}
where the last inequality follows from the data processing inequality. 
Notice that in this setting $\sK_{\mathrm{G}}^\sigma$ is an $\calS$-constrained Gaussian kernel where  
\begin{equation*}
    \mathcal{S} \coloneqq \bigcup_{B\subset[n]} \psi_{B}(\mathcal{W}).
\end{equation*}
It follows from Lemma~\ref{Lemma:ProjectedDPSGDDomain} that $\mathrm{dia}(\mathcal{S}) \leq D + 2\eta C$. Therefore, Proposition~\ref{Proposition:ContractionCoefficientGaussianKernel} implies that
\begin{equation*}
    \sE_\eps\big(\sK_{\mathrm{G}}^{\sigma}(\psi_{B}\mu) \Vert \sK_{\mathrm{G}}^{\sigma}(\psi_{B}'\nu)\big) \leq \theta_\eps\bigg(\frac{D + 2\eta C}{\sigma}\bigg) \sE_\eps(\psi_{B}\mu \Vert \psi'_{B}\nu),
\end{equation*}
and, as a result,
\begin{equation*}
    \sE_\eps(\sK\mu \Vert \sK'\nu) \leq \sum_{B\subset[n]} p^{\lvert B \rvert} (1-p)^{n-\lvert B \rvert} \theta \sE_\eps(\psi_{B}\mu \Vert \psi_{B}'\nu),
\end{equation*}
where $\theta = \theta_\eps\big((D+2\eta C)/\sigma\big)$. A straightforward manipulation shows that
\begin{align*}
    \sE_\eps(\sK\mu \Vert \sK'\nu) &\leq \sum_{B\subset[n]} \mathbf{1}_{\{n\notin B\}} p^{\lvert B \rvert} (1-p)^{n-\lvert B \rvert} \theta \sE_\eps(\psi_{B}\mu \Vert \psi_{B}'\nu)\\
    &\quad + \sum_{B\subset[n]} \mathbf{1}_{\{n\in B\}} p^{\lvert B \rvert} (1-p)^{n-\lvert B \rvert} \theta \sE_\eps(\psi_{B}\mu \Vert \psi_{B}'\nu).
\end{align*}
By the definition of $\psi_{B}$ in \eqref{eq:DefPsiBProjected}, we note that $\psi_{B} = \psi_{B}'$ whenever $n \notin B$. Hence, the data processing inequality implies that
\begin{align*}
    \sE_\eps(\sK\mu \Vert \sK'\nu) &\leq \sum_{B\subset[n]} \mathbf{1}_{\{n\notin B\}} p^{\lvert B \rvert} (1-p)^{n-\lvert B \rvert} \theta \sE_\eps(\mu \Vert \nu)\\
    &\quad + \sum_{B\subset[n]} \mathbf{1}_{\{n\in B\}} p^{\lvert B \rvert} (1-p)^{n-\lvert B \rvert} \theta \sE_\eps(\psi_{B}\mu \Vert \psi_{B}'\nu).
\end{align*}
A simple combinatorial argument implies that
\begin{equation*}
    \sE_\eps(\sK\mu \Vert \sK'\nu) \leq (1-p) \theta \sE_\eps(\mu \Vert \nu) + p\theta,
\end{equation*}
where we used the trivial upper bound $\sE_\eps(\psi_{B}\mu \Vert \psi_{B}'\nu) \leq 1$.
\end{proof}

\begin{proof}[\textbf{Proof of Proposition~\ref{Proposition:CoupledNonlinearDPIProjectedSWR}}]
Let $\mathcal{X} = \{x_{1},\ldots,x_{n}\}$ and $\mathcal{X}' = \{x'_{1},\ldots,x'_{n}\}$. Without loss of generality, we may assume that $x_{n} \neq x'_{n}$ and $x_{i} = x_{i}'$ for $i\in[n-1]$. Recall the definition of $\Psi$ in \eqref{eq:DefPsiSampleWithoutReplacement}. In particular, we have that
\begin{equation*}
    \sK = \frac{1}{\binom{n}{b}} \sum_{B\subset[n] : \lvert B \rvert = b} \Pi \circ \sK_{\mathrm{G}}^{\sigma} \circ \psi_{B} \qquad \text{and} \qquad \sK' = \frac{1}{\binom{n}{b}} \sum_{B\subset[n] : \lvert B \rvert = b} \Pi \circ \sK_{\mathrm{G}}^{\sigma} \circ \psi_{B}'.
\end{equation*}
By the  convexity of $(P,Q) \mapsto \sE_\eps(P \Vert Q)$, the previous expressions lead to
\begin{align*}
    \sE_\eps(\sK\mu \Vert \sK'\nu) &\leq \frac{1}{\binom{n}{b}} \sum_{B\subset[n] : \lvert B \rvert = b} \sE_\eps\big((\Pi \circ \sK_{\mathrm{G}}^{\sigma} \circ \psi_{B}) \mu \Vert (\Pi \circ \sK_{\mathrm{G}}^{\sigma} \circ \psi_{B}') \nu\big)\\
    &= \frac{1}{\binom{n}{b}} \sum_{B\subset[n] : \lvert B \rvert = b} \sE_\eps\big((\Pi \circ \sK_{\mathrm{G}}^{\sigma})(\psi_{B}\mu) \Vert (\Pi \circ \sK_{\mathrm{G}}^{\sigma})(\psi_{B}'\nu)\big)\\
    &\leq \frac{1}{\binom{n}{b}} \sum_{B\subset[n] : \lvert B \rvert = b} \sE_\eps\big(\sK_{\mathrm{G}}^{\sigma}(\psi_{B}\mu) \Vert \sK_{\mathrm{G}}^{\sigma}(\psi_{B}'\nu)\big),
\end{align*}
where the last inequality follows from the data processing inequality. As in the proof of Proposition~\ref{Proposition:CoupledNonlinearDPIProjectedPoisson}, $\sK_{\mathrm{G}}^{\sigma}$ is an $\calS$-constrained Gaussian kernel where 
\begin{equation*}
    \mathcal{S} \coloneqq \bigcup_{B\subset[n] : \lvert B \rvert = b} \psi_{B}(\mathcal{W}).
\end{equation*}
In light of Lemma~\ref{Lemma:ProjectedDPSGDDomain}, we have $\mathrm{dia}(\mathcal{S}) \leq D + 2\eta C$. Therefore, Proposition~\ref{Proposition:ContractionCoefficientGaussianKernel} implies that
\begin{equation*}
    \sE_\eps\big(\sK_{\mathrm{G}}^{\sigma}(\psi_{B}\mu) \Vert \sK_{\mathrm{G}}^{\sigma}(\psi_{B}'\nu)\big) \leq \theta_\eps\bigg(\frac{D + 2\eta C}{\sigma}\bigg) \sE_\eps(\psi_{B}\mu \Vert \psi'_{B}\nu),
\end{equation*}
and, as a result,
\begin{equation*}
    \sE_\eps(\sK\mu \Vert \sK'\nu) \leq \frac{1}{\binom{n}{b}} \sum_{B\subset[n] : \lvert B \rvert = b} \theta \sE_\eps(\psi_{B}\mu \Vert \psi_{B}'\nu),
\end{equation*}
where $\theta = \theta_\eps\big((D+2\eta C)/\sigma\big)$. A straightforward manipulation shows that
\begin{align*}
    \sE_\eps(\sK\mu \Vert \sK'\nu) &\leq \frac{1}{\binom{n}{b}} \sum_{B\subset[n] : \lvert B \rvert = b} \mathbf{1}_{\{n\notin B\}} \theta \sE_\eps(\psi_{B}\mu \Vert \psi_{B}'\nu)\\
    &\quad + \frac{1}{\binom{n}{b}} \sum_{B\subset[n] : \lvert B \rvert = b} \mathbf{1}_{\{n\in B\}} \theta \sE_\eps(\psi_{B}\mu \Vert \psi_{B}'\nu).
\end{align*}
By the definition $\psi_{B}$ in \eqref{eq:DefPsiBProjected}, we note that $\psi_{B} = \psi_{B}'$ whenever $n \notin B$. Hence, the data processing inequality implies that
\begin{equation*}
    \sE_\eps(\sK\mu \Vert \sK'\nu) \leq \frac{1}{\binom{n}{b}} \sum_{B\subset[n] : \lvert B \rvert = b} \mathbf{1}_{\{n\notin B\}} \theta \sE_\eps(\mu \Vert \nu) + \frac{1}{\binom{n}{b}} \sum_{B\subset[n] : \lvert B \rvert = b} \mathbf{1}_{\{n\in B\}} \theta \sE_\eps(\psi_{B}\mu \Vert \psi_{B}'\nu).
\end{equation*}
A simple combinatorial argument implies that
\begin{equation*}
    \sE_\eps(\sK\mu \Vert \sK'\nu) \leq \bigg(1-\frac{b}{n}\bigg) \theta \sE_\eps(\mu \Vert \nu) + \frac{b}{n}\theta,
\end{equation*}
where we used the trivial upper bound $\sE_\eps(\psi_{B}\mu \Vert \psi_{B}'\nu) \leq 1$.
\end{proof}

\begin{proof}[\textbf{Proof of Proposition~\ref{Prop:RegularizedCouplingControl}}]
We proceed by induction on $t\geq0$. Since $\mu_{0} = \tilde{\mu}_{0}$ by assumption, the desired inequality holds with equality for $t=0$. Assume that $\displaystyle \lVert \tilde{\mu}_{t} - \mu_{t} \rVert_{\mathrm{TV}} \leq 1-\mathbb{P}(\chi^{2}_{d} \leq \kappa^{2})^{t}$ for some $t\in\mathbb{N}$. Since $\tilde{\mu}_{t+1} = \tilde{\sK}_{t+1} \tilde{\mu}_{t}$ and $\tilde{\sK}_{t+1} = \sK_{\mathrm{G}}^{\sigma} \circ \Psi$, we obtain that 
\begin{equation*}
    \tilde{\mu}_{t+1} = (\sK_{\mathrm{G}}^{\sigma} \circ \Psi) \tilde{\mu}_{t} = \sK_{\mathrm{G}}^{\sigma}(\Psi\tilde{\mu}_{t}).
\end{equation*}
By the definition of the Gaussian kernel $\sK_{\mathrm{G}}^{\sigma}$, we have that
\begin{equation*}
    \phantom{\forall \nu\in\mathcal{P}(\mathbb{R}^{d}).} \qquad \qquad \qquad \sK_{\mathrm{G}}^{\sigma} \nu = \int_{\mathbb{R}^{d}} \nu \ast \delta_{\sigma x} \mathrm{d}G(x), \qquad \qquad \qquad \forall \nu\in\mathcal{P}(\mathbb{R}^{d}).
\end{equation*}
where $G$ is the standard Gaussian distribution in $\mathbb{R}^{d}$ and $\ast$ is the additive convolution operator in $\mathcal{P}(\mathbb{R}^{d})$. In particular, we obtain that
\begin{align}
    \nonumber \tilde{\mu}_{t+1} &= \int_{\mathbb{R}^{d}} \Psi \tilde{\mu}_{t} \ast \delta_{\sigma x} \mathrm{d}G(x)\\
    \label{eq:RegularizedCouplingControlTildeMu} &= \int_{\mathbb{B}_{d}(\kappa)} \Psi \tilde{\mu}_{t} \ast \delta_{\sigma x} \mathrm{d}G(x) + \int_{\mathbb{B}_{d}(\kappa)^{c}} \Psi \tilde{\mu}_{t} \ast \delta_{\sigma x} \mathrm{d}G(x),
\end{align}
where $\mathbb{B}_{d}(\kappa) \coloneqq \{w\in\mathbb{R}^d: \|w\|\leq\kappa\}$. Similarly, we can show that
\begin{equation*}
    \mu_{t+1} = \Pi_{t+1}\bigg(\int_{\mathbb{R}^{d}} \Psi \mu_{t} \ast \delta_{\sigma x} \mathrm{d}G(x)\bigg).
\end{equation*}
Since the kernel $\Pi_{t+1}$ is linear over $\mathcal{P}(\mathbb{R}^{d})$, we obtain that
\begin{align}
    \nonumber \mu_{t+1} &= \int_{\mathbb{R}^{d}} \Pi_{t+1}(\Psi \mu_{t} \ast \delta_{\sigma x}) \mathrm{d}G(x)\\
    \label{eq:RegularizedCouplingControlMu} &= \int_{\mathbb{B}_{d}(\kappa)} \Pi_{t+1}(\Psi \mu_{t} \ast \delta_{\sigma x}) \mathrm{d}G(x) + \int_{\mathbb{B}_{d}(\kappa)^{c}} \Pi_{t+1}(\Psi \mu_{t} \ast \delta_{\sigma x}) \mathrm{d}G(x).
\end{align}
Recall that $\mu_{t} = \sK_{t} \mu_{t-1}$ and $\sK_{t} = \Pi_{t} \circ \sK_{\mathrm{G}}^{\sigma} \circ \Psi$ where $\Pi_{t}$ is the projection onto $\mathcal{W}_{t} = \mathbb{B}_{d}(r_{t})$. In particular, we have that $\mathrm{Supp}(\mu_{t}) \subseteq \mathbb{B}_{d}(r_{t})$. By the definition of $\psi_{B}$ in \eqref{eq:DefPsiBRegularized}, it is immediate to verify that, for all $w\in\mathbb{B}_{d}(r_{t})$ and $B\subset[n]$,
\begin{equation*}
    \lVert \psi_{B}(w) \rVert \leq (1-\lambda)r_{t} + \eta C.
\end{equation*}
As a result, we obtain that $\mathrm{Supp}(\Psi\mu_{t}) \subseteq \mathbb{B}_{d}\big((1-\lambda)r_{t}+\eta C\big)$. By the triangle inequality, for any $w\in\mathbb{B}_{d}\big((1-\lambda)r_{t}+\eta C\big)$ and $x\in\mathbb{B}_{d}(\kappa)$,
\begin{equation*}
    \lVert w + \sigma x \rVert \leq (1-\lambda)r_{t} + \eta C + \sigma \kappa = r_{t+1},
\end{equation*}
where the equality follows from the definition of $r_{t}$ in \eqref{eq:RegularizedRadious}. Therefore, for all $x\in\mathbb{B}_{d}(\kappa)$, we have that $\mathrm{Supp}(\Psi\mu_{t} \ast \delta_{\sigma x}) \subseteq \mathbb{B}_{d}(r_{t+1})$ and
\begin{equation*}
    \Pi_{t+1}\big(\Psi \mu_{t} \ast \delta_{\sigma x}\big) = \Psi \mu_{t} \ast \delta_{\sigma x}.
\end{equation*}
By plugging the previous equality into \eqref{eq:RegularizedCouplingControlMu}, we obtain that
\begin{equation*}
    \mu_{t+1} = \int_{\mathbb{B}_{d}(\kappa)} \Psi \mu_{t} \ast \delta_{\sigma x} \mathrm{d}G(x) + \int_{\mathbb{B}_{d}(\kappa)^{c}} \Pi_{t+1}(\Psi \mu_{t} \ast \delta_{\sigma x}) \mathrm{d}G(x).
\end{equation*}
The previous expression and \eqref{eq:RegularizedCouplingControlTildeMu} lead to
\begin{align*}
    \lVert \tilde{\mu}_{t+1} - \mu_{t+1} \rVert_{\mathrm{TV}} &\leq \int_{\mathbb{B}_{d}(\kappa)} \lVert \Psi \tilde{\mu}_{t} \ast \delta_{\sigma x} - \Psi \mu_{t} \ast \delta_{\sigma x} \rVert_{\mathrm{TV}} \mathrm{d}G(x)\\
    & \quad + \int_{\mathbb{B}_{d}(\kappa)^{c}} \lVert \Psi \tilde{\mu}_{t} \ast \delta_{\sigma x} - \Pi_{t+1}(\Psi \mu_{t} \ast \delta_{\sigma x})\rVert_{\mathrm{TV}} \mathrm{d}G(x).
\end{align*}
By the data processing inequality, we have that
\begin{equation*}
    \lVert \Psi \tilde{\mu}_{t} \ast \delta_{\sigma x} - \Psi \mu_{t} \ast \delta_{\sigma x} \rVert_{\mathrm{TV}} \leq \lVert \tilde{\mu}_{t} - \mu_{t} \rVert_{\mathrm{TV}}.
\end{equation*}
Since the total variation distance is bounded by 1, we conclude that
\begin{align*}
    \lVert \tilde{\mu}_{t+1} - \mu_{t+1} \rVert_{\mathrm{TV}} &\leq \lVert \tilde{\mu}_{t} - \mu_{t} \rVert_{\mathrm{TV}} \int_{\mathbb{B}_{d}(\kappa)} \mathrm{d}G(x) + \int_{\mathbb{B}_{d}(\kappa)^{c}} \mathrm{d}G(x)\\
    &= \lVert \tilde{\mu}_{t} - \mu_{t} \rVert_{\mathrm{TV}} \mathbb{P}(\chi^{2}_{d} \leq \kappa^{2}) + 1 - \mathbb{P}(\chi^{2}_{d} \leq \kappa^{2}).
\end{align*}
The theorem follows from the previous inequality and the induction hypothesis.
\end{proof}

\begin{proof}[\textbf{Proof of Proposition~\ref{Proposition:RegularizedCoupledNonLinearSDPI}}]
Let $\mathcal{X} = \{x_{1},\ldots,x_{n}\}$ and $\mathcal{X}' = \{x'_{1},\ldots,x'_{n}\}$. Without loss of generality, we may assume that $x_{n} \neq x'_{n}$ and $x_{i} = x_{i}'$ for $i\in[n-1]$. Recall the definition of $\Psi$ in \eqref{eq:RegularizedDefPsi}. In particular, we have that
\begin{equation*}
    \sK_t = \frac{1}{\binom{n}{b}} \sum_{B\subset[n] : \lvert B \rvert = b} \Pi_{t} \circ \sK_{\mathrm{G}}^{\sigma} \circ \psi_{B} \qquad \text{and} \qquad \sK_t' = \frac{1}{\binom{n}{b}} \sum_{B\subset[n] : \lvert B \rvert = b} \Pi_{t} \circ \sK_{\mathrm{G}}^{\sigma} \circ \psi_{B}'.
\end{equation*}
By the convexity of $(P,Q) \mapsto \sE_\eps(P \Vert Q)$, the previous expressions lead to
\begin{align*}
    \sE_\eps(\sK_t\mu \Vert \sK_t'\nu) &\leq \frac{1}{\binom{n}{b}} \sum_{B\subset[n] : \lvert B \rvert = b} \sE_\eps\big((\Pi_{t} \circ \sK_{\mathrm{G}}^{\sigma} \circ \psi_{B}) \mu \Vert (\Pi_{t} \circ \sK_{\mathrm{G}}^{\sigma} \circ \psi_{B}') \nu\big)\\
    &= \frac{1}{\binom{n}{b}} \sum_{B\subset[n] : \lvert B \rvert = b} \sE_\eps\big((\Pi_{t} \circ \sK_{\mathrm{G}}^{\sigma}) (\psi_{B}\mu) \Vert (\Pi_{t} \circ \sK_{\mathrm{G}}^{\sigma})(\psi_{B}'\nu)\big)\\
    &\leq \frac{1}{\binom{n}{b}} \sum_{B\subset[n] : \lvert B \rvert = b} \sE_\eps\big(\sK_{\mathrm{G}}^{\sigma} (\psi_{B}\mu) \Vert \sK_{\mathrm{G}}^{\sigma} (\psi_{B}'\nu)\big),
\end{align*}
where the last inequality follows from the data processing inequality. By the definition of $\psi_{B}$ in \eqref{eq:DefPsiBRegularized}, it is immediate to verify that, for all $w\in\mathbb{B}_{d}(r_{t-1})$ and $B\subset[n]$,
\begin{equation*}
    \lVert \psi_{B}(w) \rVert \leq (1-\lambda)r_{t-1} + \eta C = r_{t} - \sigma\kappa.
\end{equation*}
As a result, we obtain that $\mathrm{Supp}(\psi_{B}\mu) \subset \mathbb{B}_{d}(r_{t}-\sigma\kappa)$. \textit{Mutatis mutandis}, we can also show that $\mathrm{Supp}(\psi_{B}'\nu) \subset \mathbb{B}_{d}(r_{t}-\sigma\kappa)$. Therefore, Proposition~\ref{Proposition:ContractionCoefficientGaussianKernel} implies that
\begin{equation*}
    \sE_\eps\big(\sK_{\mathrm{G}}^{\sigma} (\psi_{B}\mu) \Vert \sK_{\mathrm{G}}^{\sigma} (\psi_{B}'\nu)\big) \leq \theta_{\gamma}\bigg(\frac{r_{t}-\sigma\kappa}{\sigma}\bigg) \sE_\eps(\psi_{B}\mu \Vert \psi_{B}'\nu),
\end{equation*}
and, as a result,
\begin{equation*}
    \sE_\eps(\sK_t\mu \Vert \sK_t'\nu) \leq \frac{1}{\binom{n}{b}} \sum_{B\subset[n] : \lvert B \rvert = b} \theta \sE_\eps(\psi_{B}\mu \Vert \psi_{B}'\nu),
\end{equation*}
where $\theta \coloneqq \theta_\eps\big((r_{t}-\sigma\kappa)/\sigma\big)$. A straightforward manipulation shows that
\begin{align*}
    \sE_\eps(\sK_t\mu \Vert \sK_t'\nu) &\leq \frac{1}{\binom{n}{b}} \sum_{B\subset[n] : \lvert B \rvert = b} \mathbf{1}_{\{n\notin B\}} \theta \sE_\eps(\psi_{B}\mu \Vert \psi_{B}'\nu)\\
    & \quad + \frac{1}{\binom{n}{b}} \sum_{B\subset[n] : \lvert B \rvert = b} \mathbf{1}_{\{n\in B\}} \theta \sE_\eps(\psi_{B}\mu \Vert \psi_{B}'\nu).
\end{align*}
By the definition $\psi_{B}$ in \eqref{eq:DefPsiBRegularized}, we note that $\psi_{B} = \psi_{B}'$ whenever $n \notin B$. Hence, the data processing inequality implies that
\begin{equation*}
    \sE_\eps(\sK_t\mu \Vert \sK_t'\nu) \leq \frac{1}{\binom{n}{b}} \sum_{B\subset[n] : \lvert B \rvert = b} \mathbf{1}_{\{n\notin B\}} \theta \sE_\eps(\mu \Vert \nu) + \frac{1}{\binom{n}{b}} \sum_{B\subset[n] : \lvert B \rvert = b} \mathbf{1}_{\{n\in B\}} \theta \sE_\eps(\psi_{B}\mu \Vert \psi_{B}'\nu).
\end{equation*}
A simple combinatorial argument implies that
\begin{equation*}
    \sE_\eps(\sK_t\mu \Vert \sK_t'\nu) \leq \bigg(1-\frac{b}{n}\bigg) \theta \sE_\eps(\mu \Vert \nu) + \frac{b}{n} \theta,
\end{equation*}
where we used the trivial upper bound $\sE_\eps(\psi_{B}\mu \Vert \psi_{B}'\nu) \leq 1$.
\end{proof}

\begin{proof}[\textbf{Proof of Theorem~\ref{Theorem:RegularizedDPSGD}}]
Let $\mathcal{X}$ and $\mathcal{X}'$ be two neighboring datasets. Let $\tilde{\mu}_{T}$ and $\tilde{\mu}_{T}'$ be the corresponding distributions under the regularized iterative process \eqref{eq:GeneralUpdateRuleRegularized}  and $\mu_{T}$ and $\mu_{T}'$ the corresponding distributions under the projected iterative process \eqref{eq:GeneralUpdateRuleRegularizedProjected}. In \cite{liu2016e}, Liu \textit{et al.}\ established that for $\eps_{1},\eps_{2}\geq0$ and probability measures $P,Q$, and $S$,
\begin{equation}
\label{eq:LiuInequality}
    \sE_{\eps_{1}+\eps_{2}}(P \Vert Q) \leq \sE_{\eps_{1}}(P \Vert S) + e^{\eps_{1}} \sE_{\eps_{2}}(S \Vert Q).
\end{equation}
A repeated application of \eqref{eq:LiuInequality} leads to
\begin{align*}
    \sE_\eps(\tilde{\mu}_{T} \Vert \tilde{\mu}_{T}') &\leq \sE_0(\tilde{\mu}_{T} \Vert \mu_{T}) + \sE_{\eps}(\mu_{T} \Vert \tilde{\mu}_{T}')\\
    &\leq \sE_{0}(\tilde{\mu}_{T} \Vert \mu_{T}) + \sE_{\eps}(\mu_{T} \Vert \mu_{T}') + e^{\eps} \sE_{0}(\mu_{T}' \Vert \tilde{\mu}_{T}').
\end{align*}
Since $E_0(P \Vert Q) = \lVert P - Q \rVert_{\mathrm{TV}}$, Proposition~\ref{Prop:RegularizedCouplingControl} implies that
\begin{equation*}
    \sE_{\eps}(\tilde{\mu}_{T} \Vert \tilde{\mu}_{T}') \leq (1+e^\eps)\Big[1-\mathbb{P}(\chi^{2}_{d} \leq \kappa^{2})^{T}\Big] + \sE_{\eps}(\mu_{T} \Vert \mu_{T}').
\end{equation*}
The theorem follows from Proposition~\ref{Proposition:RegularizedCoupledNonLinearSDPI} and Lemma~\ref{Lemma:HSDivergenceRecursion}.
\end{proof}
%%%%%%%

%\end{comment}

\normalsize
\bibliography{references}
\bibliographystyle{alpha}

\end{document}